\documentclass[conference]{IEEEtran}
\IEEEoverridecommandlockouts
\usepackage{cite}
\usepackage{amsmath,amssymb,amsfonts}
\usepackage{algorithmic}
\usepackage{graphicx}
\usepackage{textcomp}
\usepackage{xcolor}
\usepackage{makecell}
\def\BibTeX{{\rm B\kern-.05em{\sc i\kern-.025em b}\kern-.08em
    T\kern-.1667em\lower.7ex\hbox{E}\kern-.125emX}}
\begin{document}

\title{Low-Altitude Channel Multipath Prediction via Panoramic Perception and Vision-Language Model\\
}

\author{
    \IEEEauthorblockN{Zihang~Zeng,~Shu~Sun,~Meixia~Tao,~Zhiyong~Chen,~Jianhua~Mo,~Xiangwen~Gu}
    \IEEEauthorblockA{
    School of Information Science and Electronic Engineering, Shanghai Jiao Tong University, Shanghai 200240, China
    }
    \IEEEauthorblockA{
    Corresponding authors: Shu Sun and Meixia Tao (Emails: \{shusun, mxtao\}@sjtu.edu.cn)
    }
}

\maketitle

\begin{abstract}
Unmanned aerial vehicle (UAV) communication is expected to support a wide range of low-altitude applications in 6G mobile networks. However, traditional statistical channel models provide limited accuracy in specific environments, while deterministic methods such as ray tracing usually rely on accurate three-dimensional environment models and involve high computational complexity. Existing multimodal channel prediction approaches mainly focus on large-scale metrics such as path loss, and remain insufficient for modeling small-scale parameters. To address these limitations, this paper proposes PanoLAMP, a \textbf{Pano}ramic perception and vision-language model-based \textbf{L}ow-\textbf{A}ltitude \textbf{M}ultipath \textbf{P}rediction framework. It adopts a pretrained vision-language model as the backbone and captures the propagation environment features through panoramic RGB-D observations collected at both the transmitter and receiver to predict the delay, power, azimuth angle, and zenith angle offset relative to the line-of-sight path. Experiments are conducted on a synthetic dataset containing 18,949 UAV–vehicle links across seven UAV altitudes. Experimental results show that the proposed method consistently outperforms representative baselines in both multipath parameters and statistical metrics, and demonstrates stronger generalization across different flight heights.

\end{abstract}


\section{Introduction}

With the large-scale deployment of 5G mobile communication systems and the continuous evolution toward 6G, unmanned aerial vehicle (UAV) communication is becoming an important enabling technology for low-altitude economy~\cite{jiang2025integrated}. In urban low-altitude scenarios, building facades, rooftop edges, street canyons, the ground, and cars can introduce reflection, diffraction, scattering, and blockage, resulting in rich multipath propagation. The delay, power, azimuth angle of arrival (AoA), and zenith angle of arrival (ZoA) of multipath components jointly determine the channel impulse response, power-angle-delay profile, and beam direction. Therefore, efficient and accurate link-level multipath prediction is crucial for UAV communication system design and optimization.

Traditional channel modeling mainly follows two directions. Statistical models estimate path loss, shadow fading, cluster parameters, delay spread, and angular spread from measurement campaigns. They are compact and efficient, but only describe the average behavior of a scenario and struggle to capture the effects of the building geometry, blockage boundaries, or local scatterers associated with a specific link.
Physics-based parametric models have also been developed to balance modeling accuracy and complexity. For example, the hybrid spherical- and planar-wave model in~\cite{chen2022hybrid} combines planar-wave modeling within subarrays and spherical-wave modeling across subarrays for near-field channel characterization and spatial multiplexing analysis.
Deterministic models, especially ray tracing (RT), compute propagation paths from a three-dimensional (3D) environment map and provide clear physical interpretation. However, it depends on accurate environment models and intensive ray-intersection computation, which is difficult to maintain under fast UAV motion and a continuously varying environment.

Artificial intelligence (AI)-enabled channel modeling provides a new technical route for learning complex channel distributions from data~\cite{vasudevan2024machine}. For example, The work~\cite{ethier2024machine} shows that, beyond distance and frequency, introducing only a few scalar features, such as the total obstruction depth along the straight transmitter-receiver path, significantly improves the accuracy of machine-learning-based path loss prediction. In addition, the work~\cite{kwon2025machine} further designs diffraction parameter and morphology ratio features, which improve the path loss prediction performance of multiple machine learning models. However, existing AI-assisted channel modeling methods still suffer from insufficient environmental awareness, which limits their ability to fully characterize obstruction, reflector orientation, and local scatterer geometry around a specific link.

Recently, wireless environmental information theory~\cite{zhang2025wireless} has established a theoretical foundation for environment-aware channel modeling, showing that exploiting environmental information can reduce overhead and improve performance over the conventional statistical paradigm. Recently, an increasing number of studies have exploited multimodal information, such as RGB images, depth maps, LiDAR, and radar, to assist channel prediction~\cite{wang2024multi, Sun2025LLM4PG}. For example, MES-PLA~\cite{wang2024multi} extracts and fuses features from RGB images, point clouds, and GPS data to predict path loss, while LLM4PG~\cite{Sun2025LLM4PG} adapts a large language model (LLM) to generate path loss maps from UAV-view RGB-D images. These methods, however, mainly target large-scale channel statistics or beam-level indicators, and do not directly predict the small-scale, multipath-level parameters, such as the delay, power, and angle of arrival of each propagation path, which are essential for channel reconstruction, beamforming, and spatial resource management in UAV systems.

To address these issues, this paper proposes \textbf{Pano}ramic perception and vision-language model-based \textbf{L}ow-\textbf{A}ltitude \textbf{M}ultipath \textbf{P}rediction framework (PanoLAMP), a panoramic RGB-D and vision-language model (VLM)-based framework for low-altitude channel multipath prediction. PanoLAMP adopts a pretrained VLM as the environmental perception backbone and uses panoramic RGB-D observations at both the transmitter (Tx) and receiver (Rx) to predict the delay, power, AoA, and ZoA offsets of dominant paths relative to the geometric line-of-sight (LoS) reference. To enhance 3D spatial awareness of the VLM, depth positional encoding is introduced to inject distance priors into visual tokens. Low-rank adaptation (LoRA) is then used to efficiently adapt the pretrained VLM to the multipath prediction task. Finally, a multi-task mixture-of-experts head is designed to jointly predict multipath parameters. Experiments show that the proposed method achieves higher accuracy in multipath prediction than baselines, and maintains good generalization capability to unseen flight heights.

\begin{figure*}[htbp]
\centering
\includegraphics[width=0.9\textwidth]{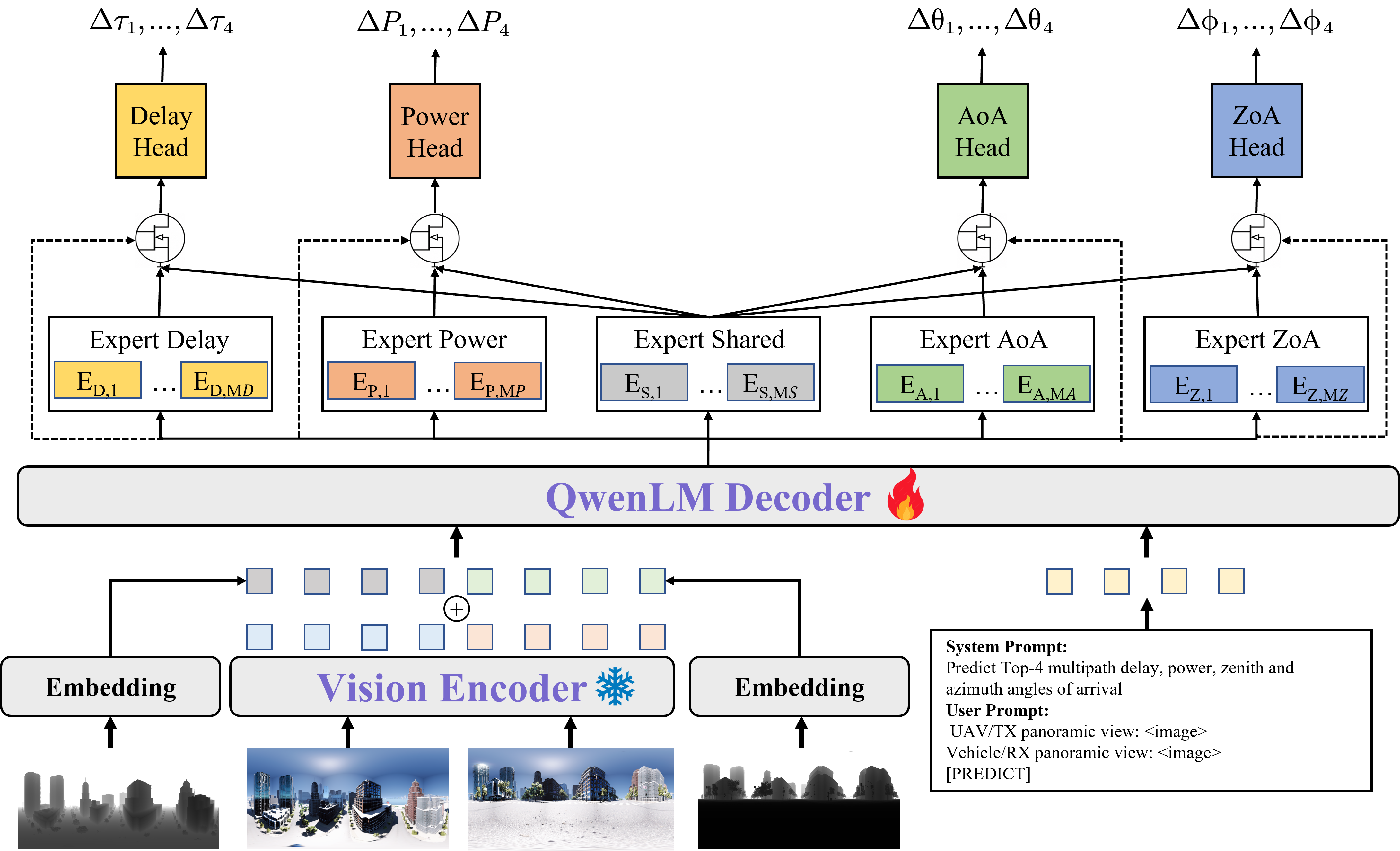}
\caption{Overview of the proposed framework. Dual-view panoramic RGB-D observations are processed by a depth-aware semantic encoder with depth positional encoding, aggregated into a $\langle\text{PREDICT}\rangle$ token by a LoRA-tuned QwenLM backbone, and decoded by a multi-task mixture-of-experts head.}
\label{fig_model}
\end{figure*}

\section{VLM-Based Multipath Channel Model}
\label{sec_method}

This section formulates the multipath prediction task, describes the proposed network, and presents the training objective.

\subsection{Problem Formulation}

Given panoramic RGB images $\mathbf{I}_{\rm tx},\mathbf{I}_{\rm rx}\in\mathbb{R}^{H\times W\times 3}$ and depth maps $\mathbf{D}_{\rm tx},\mathbf{D}_{\rm rx}\in\mathbb{R}^{H\times W}$ collected at the UAV and the car, the goal is to estimate the parameters of the top-$L$ dominant propagation paths, regardless of whether a physical LoS component exists. To improve generalisation, we represent each dominant path as an offset relative to a geometric LoS reference computed from the transmitter and receiver coordinates:
\begin{equation}
\mathbf{y}_l=\bigl(\Delta\tau_l,\Delta P_l,\Delta\theta_l,\Delta\phi_l\bigr),\quad l=1,\dots,L,
\label{eq_method_target}
\end{equation}
where $\Delta\tau_l$, $\Delta P_l$, $\Delta\theta_l$, and $\Delta\phi_l$ denote the delay, power, AoA, and ZoA offsets with respect to the reference $(\tau_0,P_0,\theta_0,\phi_0)$. The prediction task is formulated as
\begin{equation}
f_\Theta:\bigl(\mathbf{I}_{\rm tx},\mathbf{D}_{\rm tx},\mathbf{I}_{\rm rx},\mathbf{D}_{\rm rx}\bigr)\longmapsto\bigl\{\Delta\tau_l,\Delta P_l,\Delta\theta_l,\Delta\phi_l\bigr\}_{l=1}^{L},
\label{eq_problem_def}
\end{equation}
where $\Theta$ denotes the trainable parameters. The absolute path parameters can be recovered by adding the predicted offsets to the geometric LoS reference.

\subsection{Overall Architecture}

As shown in Fig.~\ref{fig_model}, the proposed framework maps multimodal observations to path-level channel parameters through three modules. The first module is a depth-aware dual-view encoder, which converts each RGB panorama into visual tokens and adds the corresponding depth map as a continuous positional prior. In this way, the model receives 3D spatial information without increasing the token length. The depth-enhanced tokens from the UAV and car views are then concatenated with the text prompt and passed to a LoRA-adapted QwenLM backbone. A special $\langle\text{PREDICT}\rangle$ token is used to collect propagation-related information from the two views into a compact representation. The final module is a multi-task mixture-of-experts head, which decodes this representation into the multipath offsets of the dominant paths. This design allows the prediction tasks to share common propagation features while retaining task-specific branches for different tasks.

\subsection{Depth-Aware Dual-View Encoding}

The visual encoder in a VLM is usually pretrained on RGB images, and its input format and feature distribution are therefore shaped mainly by color and texture rather than metric depth. Treating the depth map as another image input not only double the number of visual tokens but also disturbs the pretrained visual feature space. To avoid these issues, the proposed encoder uses depth as a continuous positional prior and adds it to the visual tokens. This supplies 3D spatial information without increasing the sequence length or changing the basic RGB feature representation. For a view $v\in\{\rm tx,rx\}$, the RGB image is first encoded as
\begin{equation}
\mathbf{F}^{v}_{\rm rgb}=\mathrm{Enc}(\mathbf{I}_v)\in\mathbb{R}^{N\times d},
\label{eq_rgb_encoder}
\end{equation}
where $N$ is the number of visual tokens and $d$ is the feature dimension. The depth map is aligned with the token grid by adaptive average pooling,
\begin{equation}
\mathbf{D}'_v=\mathrm{AvgPool}(\mathbf{D}_v)\in\mathbb{R}^{H_p\times W_p},\quad H_pW_p=N,
\label{eq_depth_pool}
\end{equation}
where $D'_v(i,j)$ gives the mean depth within the area of token $(i,j)$. A sinusoidal depth positional encoding is then defined for $t=0,\dots,d/2-1$ as
\begin{equation}
\begin{aligned}
E^v_{\rm dep}(i,j,2t) &= \sin(\omega_t),\\
E^v_{\rm dep}(i,j,2t+1) &= \cos(\omega_t),
\end{aligned}
\label{eq_depth_pe}
\end{equation}
where $\omega_t=\alpha_{\rm dep}D'_v(i,j)/10000^{2t/d}$ and $\alpha_{\rm dep}$ scales the depth value into the encoding phase. After flattening, the depth embedding $\mathbf{E}^v_{\rm dep}\in\mathbb{R}^{N\times d}$ is added to the RGB tokens,
\begin{equation}
\mathbf{F}^v_{\rm out}=\mathbf{F}^v_{\rm rgb}+\mathbf{E}^v_{\rm dep},\quad v\in\{\rm tx,rx\}.
\label{eq_depth_injection}
\end{equation}
The two depth-enhanced views are concatenated into
\begin{equation}
\mathbf{F}_{\rm vis}=[\mathbf{F}^{\rm tx}_{\rm out};\mathbf{F}^{\rm rx}_{\rm out}]\in\mathbb{R}^{2N\times d},
\label{eq_dual_view_concat}
\end{equation}
which serves as the visual input to the backbone. The two viewpoints provide complementary information, which captures the overall building layout and rooftop structures, near-field blockers and reflectors between Tx and Rx.

\subsection{QwenLM Backbone with LoRA}
The depth-enhanced visual tokens and structured text prompts are provided together to the QwenLM backbone. A special token $\langle \text{PREDICT}\rangle$ is appended to the end of the input sequence to collect propagation-related information from the multimodal context. The hidden state of this token serves as a compact representation of the propagation environment described by the dual-view RGB-D observations.

We use LoRA~\cite{hu2022lora} for parameter-efficient adaptation. LoRA adds trainable low-rank branches only to the query, key, value, and output projection matrices in the attention layers, while keeping the original pretrained weights frozen. For any linear projection matrix $\mathbf{W}\in\mathbb{R}^{d\times d}$, the LoRA update is formulated as
\begin{equation}
\mathbf{W}'=\mathbf{W}+\Delta\mathbf{W}
=\mathbf{W}+\frac{\alpha_{\rm L}}{r}\mathbf{B}\mathbf{A},
\label{eq_lora}
\end{equation}
where $\mathbf{A}\in\mathbb{R}^{r\times d}$ and $\mathbf{B}\in\mathbb{R}^{d\times r}$ are trainable low-rank matrices, $r$ denotes the LoRA rank, and $\alpha_{\rm L}$ is the scaling factor.

\subsection{Multi-Task Mixture-of-Experts Head}

The hidden state $\mathbf{h}\in\mathbb{R}^{d}$ at the $\langle\text{PREDICT}\rangle$ position is used to predict the relative delay, power, AoA, and ZoA of dominant paths. The output head is based on a mixture-of-experts (MoE) \cite{ma2018modeling} design that includes both shared and task-specific experts. For task $k\in{\tau,P,\theta,\phi}$, the available experts consist of the shared set $\mathcal{E}_s=\{E_s^{(i)}\}_{i=1}^{N_s}$ and the task-specific set $\mathcal{E}_{t,k}=\{E_{t,k}^{(j)}\}_{j=1}^{N_t}$. The task representation is obtained by a gated combination,
\begin{equation}
\tilde{\mathbf{h}}_k=\sum_{E\in\mathcal{E}_s\cup\mathcal{E}_{t,k}}g_k^{(E)}E(\mathbf{h}),
\label{eq_ple}
\end{equation}
\begin{equation}
    \mathbf{g}_k=\mathrm{softmax}(\mathbf{W}_k\mathbf{h}),
\end{equation}
where $\mathbf{W}_k$ is the gate for task $k$, and the softmax weights $\mathbf{g}_k$ balance the shared and task-specific experts for each sample. The resulting representation is then passed through an independent prediction tower,
\begin{equation}
\hat{\mathbf{y}}_{\tau}=H_{\tau}(\tilde{\mathbf{h}}_{\tau}),\;
\hat{\mathbf{y}}_{P}=H_{P}(\tilde{\mathbf{h}}_{P}),\;
\hat{\mathbf{y}}_{\theta}=H_{\theta}(\tilde{\mathbf{h}}_{\theta}),\;
\hat{\mathbf{y}}_{\phi}=H_{\phi}(\tilde{\mathbf{h}}_{\phi}),
\label{eq_task_towers}
\end{equation}
where $\hat{\mathbf{y}}_{\tau},\hat{\mathbf{y}}_{P},\hat{\mathbf{y}}_{\phi}\in\mathbb{R}^{L}$ denote the delay, power, and ZoA offset of the top-$L$ dominant paths, respectively. The AoA offset $\hat{\mathbf{y}}_{\theta}\in\mathbb{R}^{2L}$ uses one sine-cosine pair for each path.

\subsection{Training Loss}

The model is trained with a multi-task objective,
\begin{equation}
\mathcal{L}=w_\tau\mathcal{L}_{\rm delay}+w_P\mathcal{L}_{\rm power}+w_{\rm AoA}\mathcal{L}_{\rm AoA}+w_{\rm ZoA}\mathcal{L}_{\rm ZoA},
\label{eq_total_loss}
\end{equation}
where $w_\tau, w_P, w_{\rm AoA}, w_{\rm ZoA}$ are weights that balance different loss. The delay loss uses the Huber loss, which is less sensitive to occasional distant paths with large delay offsets. The power and ZoA loss are computed by the mean squared error (MSE) in the dB and radian domains, respectively. For AoA, the MSE is applied to the sine and cosine components. Since the multipath components are unordered, the predicted paths are first matched to the ground-truth paths using the Hungarian algorithm at each training step, and the loss is then computed on the matched pairs.

\section{Dataset and Simulation Results}
\label{sec_exp}

\subsection{Dataset Construction}
\begin{figure}[!t]
\centering
\includegraphics[width=1.0\columnwidth]{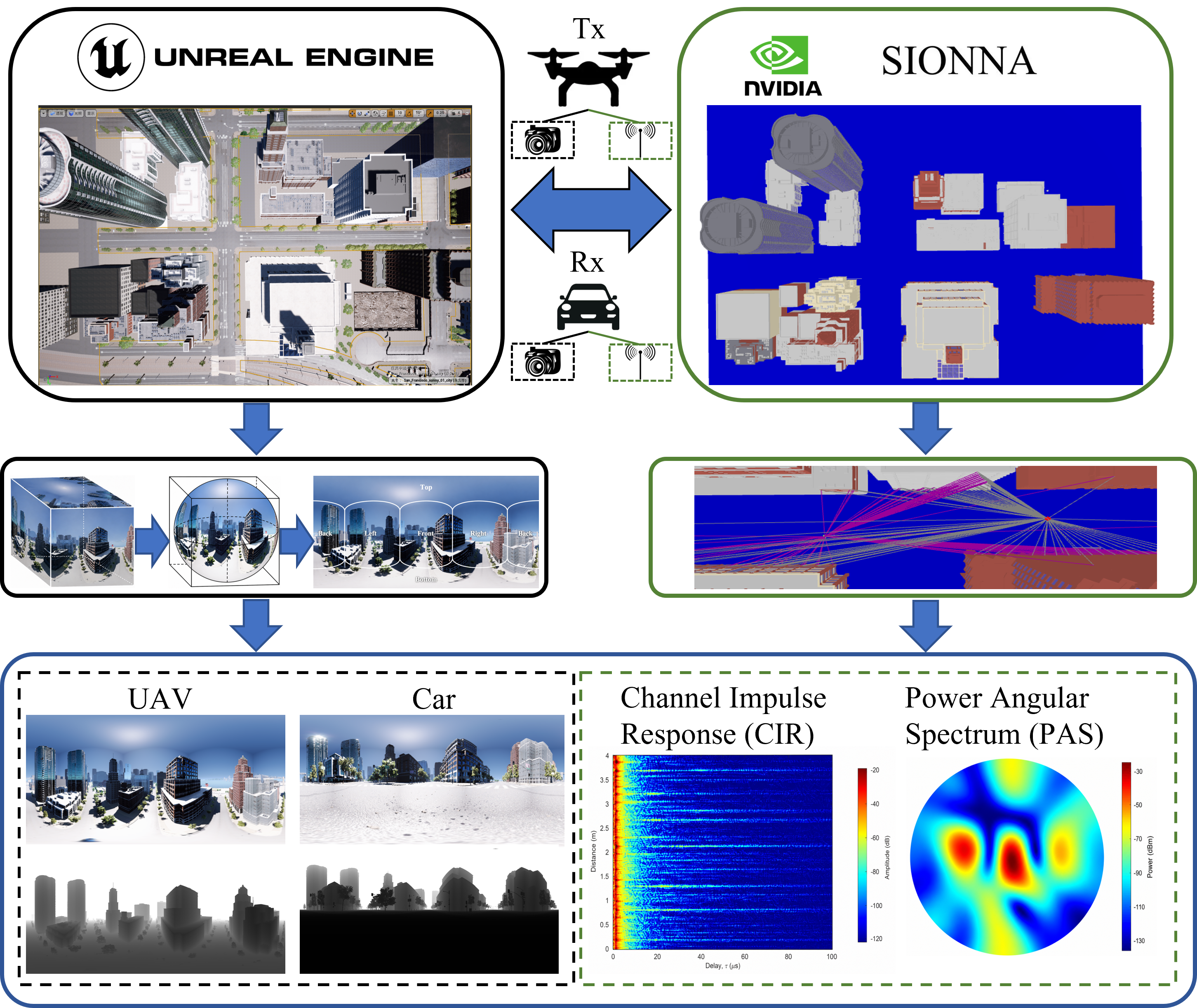}
\caption{Multimodal dataset construction pipeline.}
\label{fig_dataset}
\end{figure}

The multimodal dataset is constructed by combining AirSim and Sionna RT, as shown in Fig.~\ref{fig_dataset}. AirSim~\cite{shah2017airsim}, developed by Microsoft, provides high-fidelity simulation for UAVs and other autonomous agents, including physical motion, sensor observations, and environmental noise. It is therefore suitable for generating UAV perception data. In AirSim, we build a 3D urban scene with buildings of different sizes and heights, road intersections, and cars. Wireless propagation in the same scene is simulated using Sionna RT~\cite{hoydis2023sionna}, a RT-based propagation simulator that accounts for blockage, reflection, diffraction, and scattering and is commonly used for channel modeling. To ensure consistency between visual and wireless data, the urban model built in AirSim is imported into Sionna RT with its geometry, position, and material properties preserved. A shared world coordinate system is also established across the two platforms. Since the UAV camera and the transmitting antenna are calibrated in the same coordinate frame, each perception sample and its channel label correspond to the same scene, time instant, and UAV pose.

\begin{table}[!t] \caption{Simulation Configuration} \label{tab_dataset} \centering  \setlength{\tabcolsep}{3pt} \renewcommand{\arraystretch}{1.12} \begin{tabular}{p{0.24\columnwidth}p{0.21\columnwidth}p{0.29\columnwidth}p{0.17\columnwidth}} \hline \multicolumn{2}{c}{AirSim Configuration} & \multicolumn{2}{c}{Sionna RT Configuration} \\ \hline Parameter & Value & Parameter & Value \\ \hline Modalities & RGB, depth map & Carrier frequency & 1.4\,GHz \\ Resolution & $512 \times 1024$ & Tx power & 40\,dBm \\ Depth range & 200\,m & Tx/Rx antenna pattern & Isotropic \\ UAV altitude & 30, 35, 40, 45, 50, 55, 60\,m & Number of paths & 4 \\ Car height & 1.5\,m & Propagation mechanisms & Reflection, diffraction \\ \hline \end{tabular} \end{table}

Each RGB image and depth map is synthesized from six cubemap views facing the front, back, left, right, up, and down directions. Each view covers a $90^\circ$ field of view, and the six non-overlapping views together form a full panorama. The cubemap images $I_{c}^{H_{c}\times W_{c}}$ are projected onto an equirectangular image $I_{e}^{H_{e}\times W_{e}}$ through spherical projection, where $H_{c}=W_{c}=256$, $H_{e}=512$, and $W_{e}=1024$. This projection maps the local observations on the cube faces to a continuous spherical coordinate space, thereby reducing seams between adjacent views.

For each UAV-car pair, Sionna RT computes the $L$ strongest multipath components and uses them as the channel label for the corresponding perception sample. Since reflection and diffraction are the main propagation mechanisms on urban roads, only these two interactions are enabled. All building and road surfaces are assigned concrete material properties to avoid material-dependent variations and let the model focus on the multipath propagation process. For the $l$-th path, we record four parameters: the power $P_l$, the delay $\tau_l$, the azimuth angle $\theta_l$, and the zenith angle $\phi_l$. The number of paths is set to $L=4$, because the four strongest components typically account for about $90\%$ of the total received power in outdoor environments. Each link is therefore represented by $4\times4$ channel parameters. The detailed simulation settings are listed in Table~\ref{tab_dataset}.

The urban scene is a crossroads formed by two perpendicular streets. The UAV is sampled every $10$\,m along both street arms. At each horizontal position, it is placed at seven altitudes from $30$ to $60$\,m with a $5$\,m interval, forming a layered transmitter grid in altitude. The car moves along the same streets at a fixed height of $1.5$\,m and is sampled at intervals of about $5$\,m. Each link is formed by pairing one UAV position with one car position, covering both LoS cases and non-line-of-sight (NLoS) cases caused by building blockage. The resulting dataset contains $18{,}949$ independent UAV-car links, with $2{,}707$ links at each of the seven altitudes. Each link includes four panoramic images, resulting in $75{,}796$ images in total. Each link is also annotated with four dominant multipath components, and each component contains delay, power, AoA, and ZoA values. These labels describe the channel in the temporal, energy, and spatial domains. In total, the dataset provides $303{,}184$ multipath-parameter labels for learning the mapping from multimodal perception to wireless propagation characteristics.

\subsection{Experimental Setup}

The proposed model is trained end to end using the AdamW optimizer. We use separate learning rates of $1\times10^{-4}$ for the LoRA branches and $3\times10^{-4}$ for the prediction head. Training is conducted for $100$ epochs in bf16 precision on eight NVIDIA RTX 4090 GPUs, with a per-GPU batch size of $8$ and a global batch size of $64$. The learning rate is warmed up from $1\times10^{-5}$ to $1\times10^{-4}$ over the first $10$ epochs and then annealed to $1\times10^{-7}$ using a cosine schedule. Qwen2-VL-2B is used as the backbone. The LoRA rank and scaling factor are set to $r=8$ and $\alpha_{\rm L}=16$, respectively. The mixture-of-experts head uses $N_s=2$ shared experts and $N_t=1$ task-specific expert for each task, and the depth scaling factor is set to $\alpha_{\rm dep}=100$.

\begin{figure*}[!t]
\centering
\begin{minipage}[b]{0.32\textwidth}\centering\includegraphics[width=\textwidth]{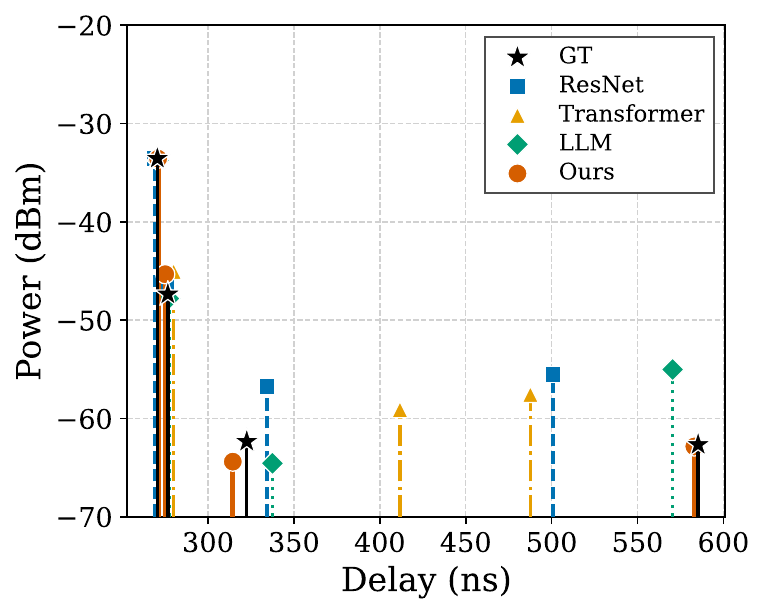}\\(a) PDP\end{minipage}
\hfill
\begin{minipage}[b]{0.33\textwidth}\centering\includegraphics[width=\textwidth]{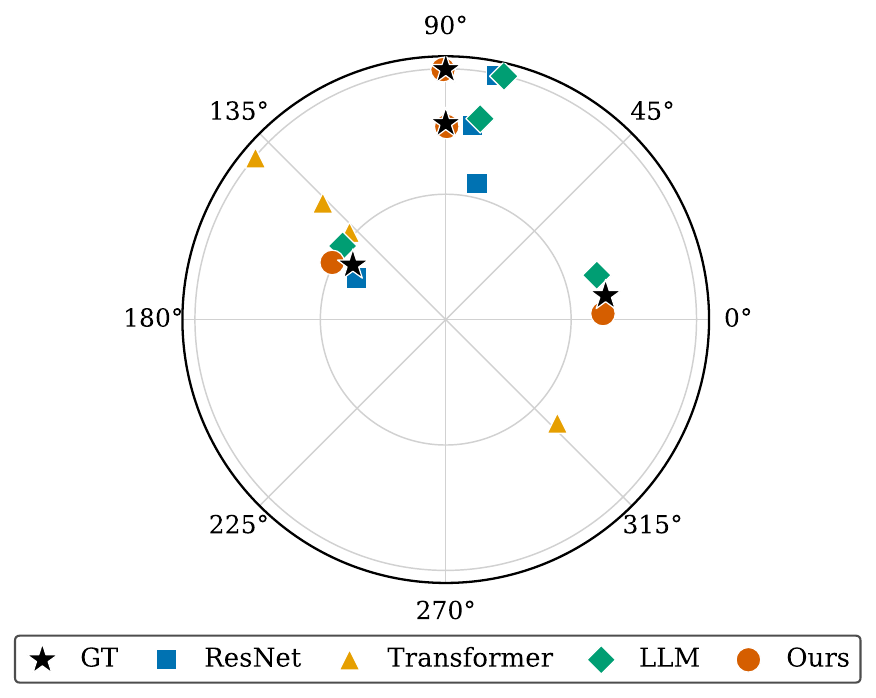}\\(b) AoA\end{minipage}
\hfill
\begin{minipage}[b]{0.33\textwidth}\centering\includegraphics[width=\textwidth]{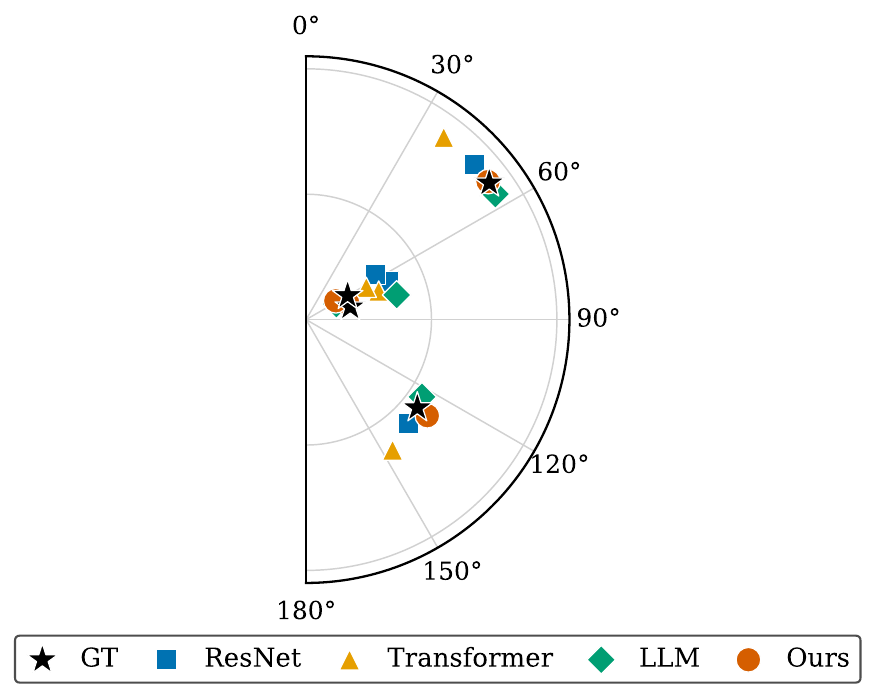}\\(c) ZoA\end{minipage}
\caption{Per-path prediction versus the ground truth on test samples: (a) PDP, (b) AoA, and (c) ZoA.}
\label{fig_samples}
\end{figure*}

\subsection{Evaluation Metrics and Baselines}

We evaluate performance at two levels. At the multipath-parameter level, the per-path delay and power are measured by the normalized mean absolute error (NMAE) and the normalized mean square error (NMSE), while the AoA and ZoA are measured by the mean cosine similarity between the predicted and ground-truth directions. We also report the mean absolute errors of the power-weighted average delay and average angle. At the channel-statistics level, the Rician $K$ factor, the root-mean-square (RMS) delay spread $\tau_{\rm rms}$, and the power-weighted angular spreads of azimuth and zenith are computed with respect to the ground truth.


We compare the proposed model with three baselines that cover different feature representations.
\begin{itemize}
\item \textbf{ResNet}: ResNet-18~\cite{he2016deep} is a convolutional neural network (CNN) built with residual blocks and shortcut connections. The residual design helps alleviate gradient vanishing and degradation in deep networks, while batch normalization improves training stability.

\item \textbf{Transformer}: Transformer~\cite{vaswani2017attention} divides an image into non-overlapping patches and linearly embeds them into a token sequence. Self-attention is then used to model long-range dependencies among image patches.

\item \textbf{LLM}: This baseline~\cite{Huang2025LLM4MG} uses a pretrained large language model for channel multipath prediction. It first extracts image tokens from different views with a frozen visual encoder, and then projects these tokens into the word embedding space of the language model. The multimodal features are finally fused through transformer blocks.
\end{itemize}

\subsection{Performance}
The samples are randomly split into training, validation, and test sets at a ratio of $80\%/10\%/10\%$. Table~\ref{tab_main} reports the multipath-level prediction accuracy. The proposed method achieves the best performance on all six metrics, with delay and power NMSE values of $0.0232$ and $0.0068$, respectively, and zenith and azimuth cosine similarities of $0.9931$ and $0.9233$. The azimuth metric shows the largest performance gap. The two vision-language methods, our method at $0.9233$ and the LLM baseline at $0.8915$, outperform ResNet at $0.7766$ and the Transformer at $0.3543$. This result suggests that scene semantics learned from large-scale pretraining are useful for predicting azimuth, which is closely related to the spatial layout of the environment. For power prediction, the NMSE of the proposed method is $0.0068$, about $0.6$ times that of the LLM baseline, indicating more accurate modeling of blockage and path-loss effects.

\begin{table}[!t]
\caption{Overall Performance Comparison.}
\label{tab_main}
\centering
\setlength{\tabcolsep}{5pt}
\begin{tabular}{lcccccc}
\hline
Method & \makecell{Delay\\NMSE} & \makecell{Delay\\NMAE} & \makecell{Power\\NMSE} & \makecell{Power\\NMAE} & \makecell{ZoA\\CosSim} & \makecell{AoA\\CosSim} \\\hline
Transformer & 0.0346 & 0.0786 & 0.0205 & 0.1021 & 0.9552 & 0.3543 \\
ResNet      & 0.0274 & 0.0615 & 0.0129 & 0.0757 & 0.9834 & 0.7766 \\
LLM         & 0.0287 & 0.0532 & 0.0107 & 0.0676 & 0.9885 & 0.8915 \\
\textbf{Ours} & \textbf{0.0232} & \textbf{0.0450} & \textbf{0.0068} & \textbf{0.0487} & \textbf{0.9931} & \textbf{0.9233} \\
\hline
\end{tabular}
\end{table}


We compare the predictions with the ground truth on representative test samples in the delay and angular domains, as shown in Fig.~\ref{fig_samples}. In the power delay profile (PDP) in Fig.~\ref{fig_samples}(a), the proposed method estimates both the delay and power of the dominant paths more accurately than the baselines, and also captures weak NLoS paths at longer delays. In the azimuth and zenith polar plots in Fig.~\ref{fig_samples}(b) and Fig.~\ref{fig_samples}(c), its predictions are the most consistent with the ground truth. This trend is particularly clear for azimuth, where the predicted directions remain close to the dominant scattering direction despite the larger angular ambiguity.

\subsection{Ablation Study}
We conduct ablation experiments on the random split to evaluate the contribution of each component, as reported in Table~\ref{tab_ablation}. For the input view, using only one view degrades all metrics. The car-only variant consistently performs better than the UAV-only variant, suggesting that the near-field environment around the receiver provides more direct propagation cues. Even so, neither single-view setting reaches the performance of dual-view fusion. For the input modality, removing depth causes only a mild performance drop, whereas removing RGB leads to a much larger degradation. This result indicates that the RGB panorama provides the main scene semantics, while depth serves as a complementary geometric cue. For the model design, removing either LoRA or the mixture-of-experts head reduces the prediction accuracy, confirming the contribution of both components.

\begin{table}[!t]
\caption{Ablation Experiment.}
\label{tab_ablation}
\centering
\footnotesize
\setlength{\tabcolsep}{1.5pt}
\begin{tabular}{lccccccc}
\hline
Variant & \makecell{$\tau_{\rm rms}$\\(ns)} & \makecell{$K$\\(dB)} & \makecell{RMS\\ZoA ($^\circ$)} & \makecell{RMS\\AoA ($^\circ$)} & \makecell{Avg.\\Delay (ns)} & \makecell{Avg.\\AoA ($^\circ$)} & \makecell{Avg.\\ZoA ($^\circ$)} \\\hline
\multicolumn{8}{l}{\textit{Input view}}\\
\quad UAV only      & 7.19 & 2.71 & 3.39 & 6.94 & 4.97 & 19.21 & 2.65 \\
\quad Car only  & 5.80 & 2.11 & 3.15 & 4.79 & 3.27 & 7.20 & 2.44 \\
\multicolumn{8}{l}{\textit{Input modality}}\\
\quad RGB only      & 5.50 & 1.99 & 2.57 & 4.38 & 3.21 & 4.93 & 2.06 \\
\quad Depth only    & 6.74 & 2.39 & 4.86 & 5.53 & 3.94 & 11.24 & 4.02 \\
\multicolumn{8}{l}{\textit{Model component}}\\
\quad $-$LoRA       & 6.32 & 2.33 & 3.93 & 5.49 & 3.70 & 10.10 & 2.98 \\
\quad $-$MoE head   & 5.60 & 2.08 & 3.09 & 4.66 & 3.19 & 6.57 & 2.74 \\
\hline
\textbf{Full (Ours)} & \textbf{4.79} & \textbf{1.68} & \textbf{2.10} & \textbf{3.79} & \textbf{2.58} & \textbf{4.82} & \textbf{1.76} \\
\hline
\end{tabular}
\end{table}

\subsection{Cross-Height Generalization}
In the cross-height split, transmitter altitudes of $30$\,m, $35$\,m, $40$\,m, $45$\,m, and $50$\,m are used for training. The altitude of $55$\,m is used for validation, and the unseen altitude of $60$\,m is used for testing. This setting evaluates how well each model generalizes to flight heights not observed during training. Table~\ref{tab_crossheight} reports the prediction accuracy of all methods under this split. The proposed method achieves the best performance across all metrics. These results suggest that the geometry-decoupled dual-view vision-language model learns scattering-structure representations from image semantics that are less tied to a particular flight altitude, leading to stronger cross-height extrapolation.

\begin{table}[!t]
\caption{Cross-height Generalization Experiment.}
\label{tab_crossheight}
\centering
\footnotesize
\setlength{\tabcolsep}{2pt}
\begin{tabular}{lccccccc}
\hline
Method & \makecell{$\tau_{\rm rms}$\\(ns)} & \makecell{$K$\\(dB)} & \makecell{RMS\\ZoA ($^\circ$)} & \makecell{RMS\\AoA ($^\circ$)} & \makecell{Avg.\\Delay (ns)} & \makecell{Avg.\\AoA ($^\circ$)} & \makecell{Avg.\\ZoA ($^\circ$)} \\\hline
Transformer & 8.85 & 3.85 & 12.51 & 8.44 & 5.08 & 74.43 & 9.64 \\
ResNet      & 10.56 & 3.55 & 11.26 & 7.79 & 6.00 & 58.91 & 5.78 \\
LLM         & 9.58 & 3.10 & 9.10 & 6.96 & 5.45 & 18.93 & 5.76 \\
\textbf{Ours} & \textbf{8.44} & \textbf{2.28} & \textbf{5.53} & \textbf{5.00} & \textbf{3.66} & \textbf{4.97} & \textbf{3.69} \\
\hline
\end{tabular}
\end{table}


\subsection{Efficiency and Complexity Evaluation}
Table~\ref{tab_efficiency} compares the model size and computational cost of different methods. The proposed method uses Qwen2-VL-2B as the backbone, resulting in about 2.22\,B total parameters, approximately 21\,TFLOPs, and a single-sample inference latency of about 143\,ms. It therefore has the largest model size and computational cost among the compared methods. With LoRA, however, only 10.35\,M parameters are trainable, which accounts for less than 0.5\% of the total parameters and remains at the same order of magnitude as the baselines. Thus, the proposed method keeps the training process parameter-efficient despite its large backbone.

\section{Conclusion}
This paper presented PanoLAMP, a low-altitude channel multipath prediction model built on panoramic perception and VLM. The proposed method outperformed the baselines at both multipath and statistical levels, and generalize well to unseen flight heights. Future work will extend the framework to real-world validation and more complex dynamic scenes, while also reducing the inference cost for real-time deployment.

\begin{table}[!t]
\caption{Efficiency and Complexity Comparison.}
\label{tab_efficiency}
\centering
\setlength{\tabcolsep}{13pt}
\renewcommand{\arraystretch}{1.12}
\begin{tabular}{lccc}
\hline
Method & \makecell{Trainable/Total \\Parameters (M)} & \makecell{FLOPs\\(G)} & \makecell{Inference\\Time (ms)} \\
\hline
Transformer & 9.17/9.2 & 0.8 & 0.93 \\
ResNet & 11.71/11.7 & 19.5 & 2.41 \\
LLM & 1.68/213.6 & 69.3 & 52.10 \\
Ours & 10.35/2220 & 21167.8 & 142.59 \\
\hline
\end{tabular}
\end{table}

\bibliographystyle{IEEEtran}
\bibliography{iccc}

\end{document}